\documentclass{article}
\usepackage{spconf,amsmath,epsfig}
\def\EUVIPPaperID{10} % *** Enter the EUVIP Paper ID here
\newcommand{\etal}{\textit{et al. }}
\usepackage[ruled,vlined]{algorithm2e}
\usepackage{algorithmic}
\usepackage{multirow}
\title{Selective Probabilistic Classifier Based on Hypothesis Testing}
\address{Paper ID \EUVIPPaperID}
\twoauthors
  {Saeed Bakhshi Germi, Esa Rahtu}
	{Tampere University\\
	Tampere, Finland}
  {Heikki Huttunen}
	{Visy Oy\\
	Tampere, Finland}
\begin{document}
\maketitle
% ----------------------------------------------------------
\begin{abstract}
In this paper, we propose a simple yet effective method to deal with the violation of the Closed-World Assumption for a classifier. Previous works tend to apply a threshold either on the classification scores or the loss function to reject the inputs that violate the assumption. However, these methods cannot achieve the low False Positive Ratio (FPR) required in safety applications. The proposed method is a rejection option based on hypothesis testing with probabilistic networks. With probabilistic networks, it is possible to estimate the distribution of outcomes instead of a single output. By utilizing Z-test over the mean and standard deviation for each class, the proposed method can estimate the statistical significance of the network certainty and reject uncertain outputs. The proposed method was experimented on with different configurations of the COCO and CIFAR datasets. The performance of the proposed method is compared with the Softmax Response, which is a known top-performing method. It is shown that the proposed method can achieve a broader range of operation and cover a lower FPR than the alternative.
\end{abstract}
% ----------------------------------------------------------
% ----------------------------------------------------------
\section{Introduction}
\footnote{“© 20XX IEEE.  Personal use of this material is permitted.  Permission from IEEE must be obtained for all other uses, in any current or future media, including reprinting/republishing this material for advertising or promotional purposes, creating new collective works, for resale or redistribution to servers or lists, or reuse of any copyrighted component of this work in other works.”}
Artificial Intelligence (AI) is becoming a vital part of many real-life applications such as healthcare, logistics, surveillance, and industry. Classification is a common concept in the AI field, and it can be considered one of the building blocks for higher-level reasoning and decision-making systems. With the increasing demand for robust and reliable algorithms, especially in safety-critical systems \cite{Aravantinos2020}, the research community has been trying to define the robustness \cite{Fawzi2017}, evaluation metrics \cite{Carlini2017}, and solutions to satisfy the requirements of a robust classifier \cite{Xu2018}. 

State-of-the-art classifiers have achieved high accuracy numbers when dealing with simple datasets such as MNIST \cite{LeCun1998} or challenging ones like ImageNet \cite{Deng2009}. However, several open questions remain on how the classifier should behave in the circumstances not covered in the training set, for example, when unseen classes appear (out-of-distribution samples) or when inputs are distorted in a way not seen in the training set. In such cases, a classifier might generate faulty results. So it becomes clear that accuracy is not enough for measuring the performance of classifiers, and the generalization to new environments and robustness to environmental changes should also be considered.

In their review, Zhang \etal argue that unexpected faulty result in a pattern recognition algorithm can happen due to the violation of either of the following assumptions\cite{Zhang2020}: (1) Closed-World Assumption where the data is assumed to have a fixed number of classes, all covered in the training set, (2) Independent and Identically Distributed Assumption where the classes in the data are assumed to be independent of each other and have the same distribution, and (3) Clean and Big Data Assumption where the data is assumed to be well-labeled and large enough for training the network properly. While fulfilling these assumptions is more accessible in a controlled environment, real-world applications rarely cover them completely.

This paper deals with the violation of the Closed-World Assumption. While a straightforward way of dealing with this issue is introducing a \textit{trash} class in the training set to cover all out-of-distribution samples, the complex distribution of them makes it impossible to train an effective classifier in most cases. Moreover, different distortions might make a sample not easy to classify, even for a human. While there is ongoing research for adversarial attacks, the phenomenon is not that common in the everyday use of AI algorithms. In a typical case, distortions usually are from these categories: blur, noise, occlusion, and digital alteration of the image.

Recent works try to solve this issue by formulating it to reliable rejection of the predictions when the network is uncertain. The rejection option, also known as selective classification, is a central concept in different classification applications when dealing with uncertainty (e.g., optical character recognition). Previous works either rely on using a specific type of activation function in the classifier, such as OpenMax \cite{Bendale2016}, temperature scaling for SoftMax \cite{Liang2017}, and Sigmoid \cite{Shu2017}, modifying the loss function such as discrepancy loss \cite{Yu2019},  using more resources such as an ensemble of multiple classifiers \cite{Lakshminarayanan2016} and Monte-Carlo dropout \cite{Gal2016}. Moreover, some also suggest a combination of different ideas \cite{Vyas2018}.

The proposed method is a rejection option based on hypothesis testing with probabilistic networks. By utilizing a Z-test over the distribution of outcomes from a probabilistic network, it is possible to estimate the statistical significance of a given output and reject insignificant results. The main difference between the proposed method and previous state-of-the-art methods such as ODIN \cite{Bendale2016} is the non-restricted use of different architectures. The proposed method can be applied to any architecture and improve the performance when dealing with violation of the Closed-World Assumption by not limiting the network to a specific loss function or activation function.

In their work, Geifman and El-Yaniv show that Softmax Response (SR) is a simple yet top-performing method in selective classifiers \cite{Geifman2017} that outperforms Monte Carlo (MC) dropout. However, this paper shows that if utilized correctly, the probabilistic network can easily outperform the SR method, making it a viable choice.

The main contributions of this paper are as follows: 
\begin{itemize}
    \item Proposing a simple yet effective method (rejection based on the statistical significance of probabilistic network output) to deal with the violation of the Closed-World Assumption in classifiers. This method can be utilized in any modern network architecture by changing the structure into a probabilistic model, which is possible with the help of existing tools.
    \item Testing the proposed method on state-of-the-art architecture (ResNet) with a diverse set of distortions (blur, noise, gamma correction, and occlusion) to show the effectiveness of the proposed method over the baseline SR method. 
\end{itemize}

The rest of this paper is structured as follows. The details of the proposed method are presented in Section~\ref{method}. Then Section~\ref{experiments} deals with the experiments and their results. Finally, Section~\ref{conclusion} concludes the work and suggests potential research directions for the future.

\begin{figure*}[!ht]
    \begin{center}
        \includegraphics[width=1\linewidth]{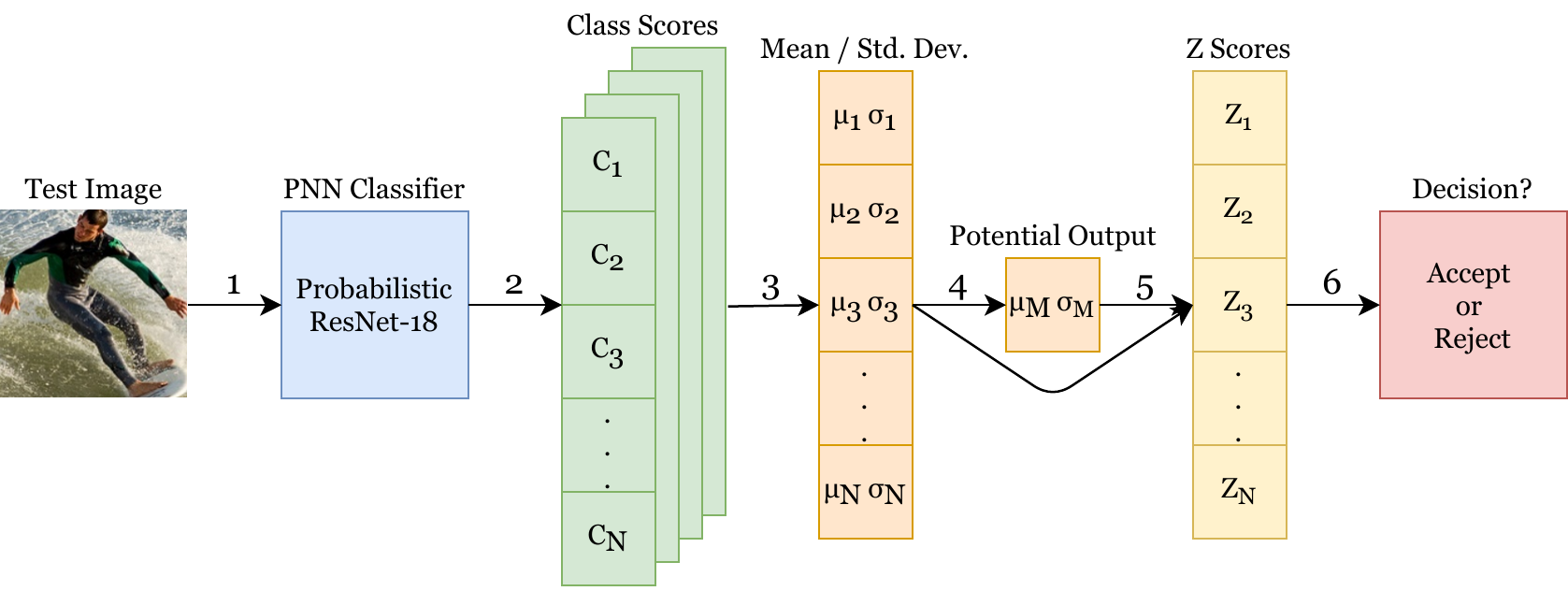}
    \end{center}
    \caption{The structure of the proposed method. (1) Pass the test image through the probabilistic classifier. (2) Repeat it $n$ times and store the class scores for each inference. (3) Calculate the mean and standard deviation for each class. (4) Find the maximum mean value and label it as potential output. (5) Run two-sample Z-tests between the potential output and all other classes, then store the Z-scores. (6) Compare Z-scores with the threshold value to decide the acceptance or rejection of the potential output.}
    \label{figure_1}
\end{figure*}

\section{Methods}
\label{method}
\subsection{Proposed method}
The proposed method requires a fully trained probabilistic classifier to work. Due to the nature of the probabilistic classifier, each inference of it will result in a slightly different class score. To utilize this fact, first, the test image is passed through the network $n$ times to get the mean and standard deviation values for each class. After that, the maximum mean value between classes is chosen as the potential output. Next, two-sample Z-tests \cite{TwoSampleZTest} are deployed between the potential output and all other classes to find the statistical significance between their difference. Finally, if the Z-scores indicate a significant difference, then the potential output is chosen to be correct. Algorithm \ref{algorithm_1} summarizes these steps and Figure \ref{figure_1} shows the structure of the proposed method.

\begin{algorithm}
    \caption{Selective Probabilistic Classifier}
    \begin{algorithmic}[1]
        \REQUIRE A trained probabilistic classifier.
            \STATE run the image through the classifier $n$ times
            \STATE find mean ($\mu$) and std. dev. ($\sigma$) for all $N$ classes
            \STATE find the class with the highest mean value ($\mathit{c}_M$)
        \FOR {$i \in 1,2,\ldots,N;\ i\neq M$} 
            \STATE run the two-sample Z-test between $\mathit{c}_M$ and $\mathit{c}_i$ 
            \STATE store the $\mathit{Z}_i$ score
        \ENDFOR
        \IF{$\mathit{Z}_i > z$ for $i \in 1,2,\ldots,C; i\neq M$} 
            \STATE set output to be $\mathit{C}_M$
        \ELSE
            \STATE set output to be Reject
        \ENDIF
    \end{algorithmic}
    \textbf{return} output value for the image
    \label{algorithm_1}
\end{algorithm}

\subsubsection{Probabilistic Neural Network}
A probabilistic neural network (PNN) classifier \cite{Mohebali2020} uses a stochastic weighting system. The classifier can allocate a class to an input sample by utilizing the posterior probability, which means each run of the network will result in a slightly different output. The amount of difference between several runs is the key to network certainty. A low standard deviation between several runs indicates a higher level of certainty for the network, making standard deviation a suitable metric for selective classification. The convolution layers for such a network are constructed based on Flipout \cite{Wen2018}. The code can be found in the Tensorflow probability directory \cite{TFP}.

\subsubsection{Two-Sample Z-test}
A Z-test \cite{Ztest} refers to any statistical test that can approximate the distribution of the hypothesis by a normal distribution. The two-sample Z-test can be used to test whether two samples are similar to each other or not. The formula is as follows:

\[\text{$Z = \frac{\mu_1-\mu_2-\Delta}{\sqrt{\frac{\sigma_1 ^ 2}{n_1} + \frac{\sigma_2 ^ 2}{n_2}}}$}\]

\noindent Where $\mu_1$ and $\mu_2$ are the mean values for two samples, $\Delta$ is the hypothesized difference between the means (0 if testing for equality), $\sigma_1$ and $\sigma_2$ are the standard deviations, and $n_1$ and $n_2$ are the sample sizes (which are equal in this paper).

By setting the null hypothesis as $H_0: \mu_1 = \mu_2$, the alternative hypothesis as $H_a: \mu_1 \neq \mu_2$, and $\Delta$ to zero, the two-sample Z-test will result in a score that indicates the likelihood of two samples being different from each other. A higher score means more likelihood for the samples to be different. This score can be compared to critical values to get the percentage for the likelihood of a significant difference between samples. These values can be found in any Z-Score table, such as \cite{ZScoreTable}.

\subsection{Softmax Response}
The SR method applies a threshold directly to the output of the Softmax layer from a deep neural network (DNN) and rejects any output below the threshold. This method was chosen as the baseline for comparison. While the method is simple, it is a known top-performer \cite{Geifman2017}.

\begin{figure*}[!ht]
    \begin{center}
        \includegraphics[width=0.9\linewidth]{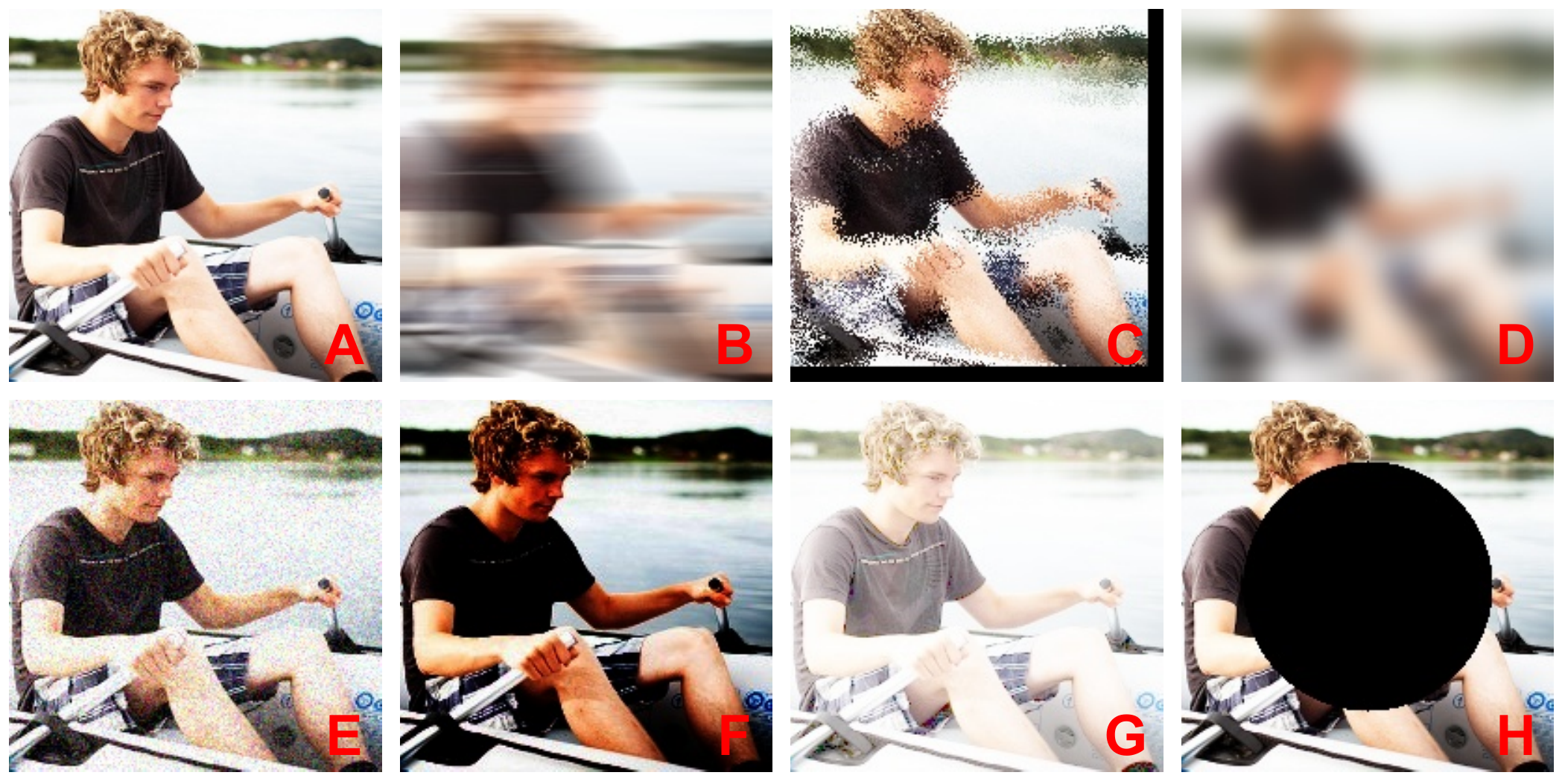}
    \end{center}
    \caption{Distortions on the image. (A) Original image. (B) Motion blur. (C) Frosted glass blur. (D) Gaussian blur. (E) Noise. (F) Gamma darkening. (G) Gamma lightening. (H) Occlusion.}
    \label{figure_2}
\end{figure*}

\section{Experiments and Results}
\label{experiments}
The proposed method was experimented on with the well-known ResNet-18 network configuration \cite{He2016}. The goal is to show the performance of it in case of violating the Closed-World Assumption. A comparison with the SR was made to evaluate the performance. This comparison was based on the area under the Receiver Operating Characteristic curve (ROC), which is threshold-independent. Both networks are trained from scratch with the same initial configuration to have a fair comparison. Other state-of-the-art methods were not included in the comparison as they either require a specific structure for the model, limiting the use case, or were only tested on more simple datasets such as MNIST.

Multiple experiments were conducted to represent various violations of Closed-World Assumption in real-world applications. In these experiments, the classifiers are trained with a limited number of classes and presented with both in-distribution and out-of-distribution samples. Further experiments also distort the test samples to see the effect of each distortion on the performance. The chosen distortions were based on \cite{Kamann2020}. Before discussing the results, the dataset and distortions are explained in detail.

\subsection{Dataset and Distortions}
\textbf{\em COCO ---}
COCO \cite{Lin2014} was chosen as the first dataset. It is a complex dataset where the objects have various sizes, qualities, and overlaps. Since the COCO is originally an object detection dataset, all instances were extracted from it manually based on the bounding boxes provided in the dataset. The data was separated into four classes: Human, Vehicle (containing 4-wheeled vehicles), Animal (containing 4-legged animals), and Background (patches of images with no overlapping objects). 260k images were used for training, excluding the animal class, and 40k images were used as test samples. The reason behind using a commonly known object detection dataset for classification is to have a more realistic dataset where an external source does not filter the samples.

\noindent\textbf{\em CIFAR ---}
CIFAR \cite{Krizhevsky2009} was chosen as the second dataset. It is a more straightforward dataset where objects are classified into ten categories. The dataset is small yet sufficiently complex, which makes it an ideal case for testing algorithms. 40k images were used for training, excluding the automobile and truck classes, and 10k images were used as test samples.

\noindent\textbf{\em Blur ---} 
Three different blurring algorithms were used to see their effect on the performance: Motion blur, Frosted glass blur, and Gaussian blur. The effect of each algorithm can be seen in Figure \ref{figure_2}(B-D). Each algorithm will simulate a situation where the object is not sharp (e.g., the camera is not focused, the object is moving, a semi-transparent object is between the camera and the object)

\noindent\textbf{\em Noise ---} 
Two different noises were added to test samples to see their effect on the performance: Gaussian noise and Salt-and-pepper noise. The effect of a sample noise can be seen in Figure \ref{figure_2}(E). It will simulate a situation where the input is noisy due to internal or external sources.

\noindent\textbf{\em Gamma Correction ---} 
The gamma correction technique was applied to each test sample to see the illumination effect on the performance. The effect of darkening and lightening can be seen in Figure \ref{figure_2}(F-G). It will simulate a situation where the amount of light in the environment changes due to environmental factors.

\noindent\textbf{\em Occlusion ---} 
A black patch was added to test samples to see the effect of occlusion on the performance. The effect of occlusion can be seen in Figure \ref{figure_2}(H). It will simulate a situation where the object is partially visible.

\begin{table*}[!ht]
    \small
    \begin{center}
        \begin{tabular}{|c||c||c||c c c|| c c||c c||c|}
            \hline
            \multirow{3}{*}{Dataset} & \multirow{3}{*}{Method} & Out & \multicolumn{3}{c||}{Blur} & \multicolumn{2}{c||}{Noise} & \multicolumn{2}{c||}{Gamma correction} & \multirow{3}{*}{Occlusion}\\
            \cline{4-10}
            & & of & \multirow{2}{*}{Motion} & Frosted & \multirow{2}{*}{Gaussian} & \multirow{2}{*}{Gaussian} & \multirow{2}{*}{S\&P} & \multirow{2}{*}{Darkening} & \multirow{2}{*}{Lightening} & \\
            & & Distribution & & glass & & & & & & \\
            \hline\hline
            \multirow{2}{*}{COCO} & Proposed & \textbf{0.65} & \textbf{0.34} & \textbf{0.25} & \textbf{0.38} & \textbf{0.22} & \textbf{0.21} & \textbf{0.16} & \textbf{0.17} & \textbf{0.23}\\
            & SR & 0.29 & 0.20 & 0.18 & 0.22 & 0.14 & 0.09 & 0.04 & 0.05 & 0.06  \\
            \hline\hline
            \multirow{2}{*}{CIFAR} & Proposed & \textbf{0.89} & \textbf{0.50} & \textbf{0.50} & \textbf{0.59} & \textbf{0.38} & \textbf{0.39} & \textbf{0.37} & \textbf{0.42} & \textbf{0.48}\\
            & SR & 0.52 & 0.44 & 0.34 & 0.47 & 0.35 & 0.25 & 0.22 & 0.26 & 0.31 \\
            \hline
        \end{tabular}
    \end{center}
    \caption{AUROC values of the tests. The values are calculated by taking the area under the ROC where the algorithm could produce a valid response.}
    \label{table_1}
\end{table*}

\subsection{Results}
\label{results}
\begin{figure}[!ht]
    \begin{center}
        \includegraphics[width=0.9\linewidth]{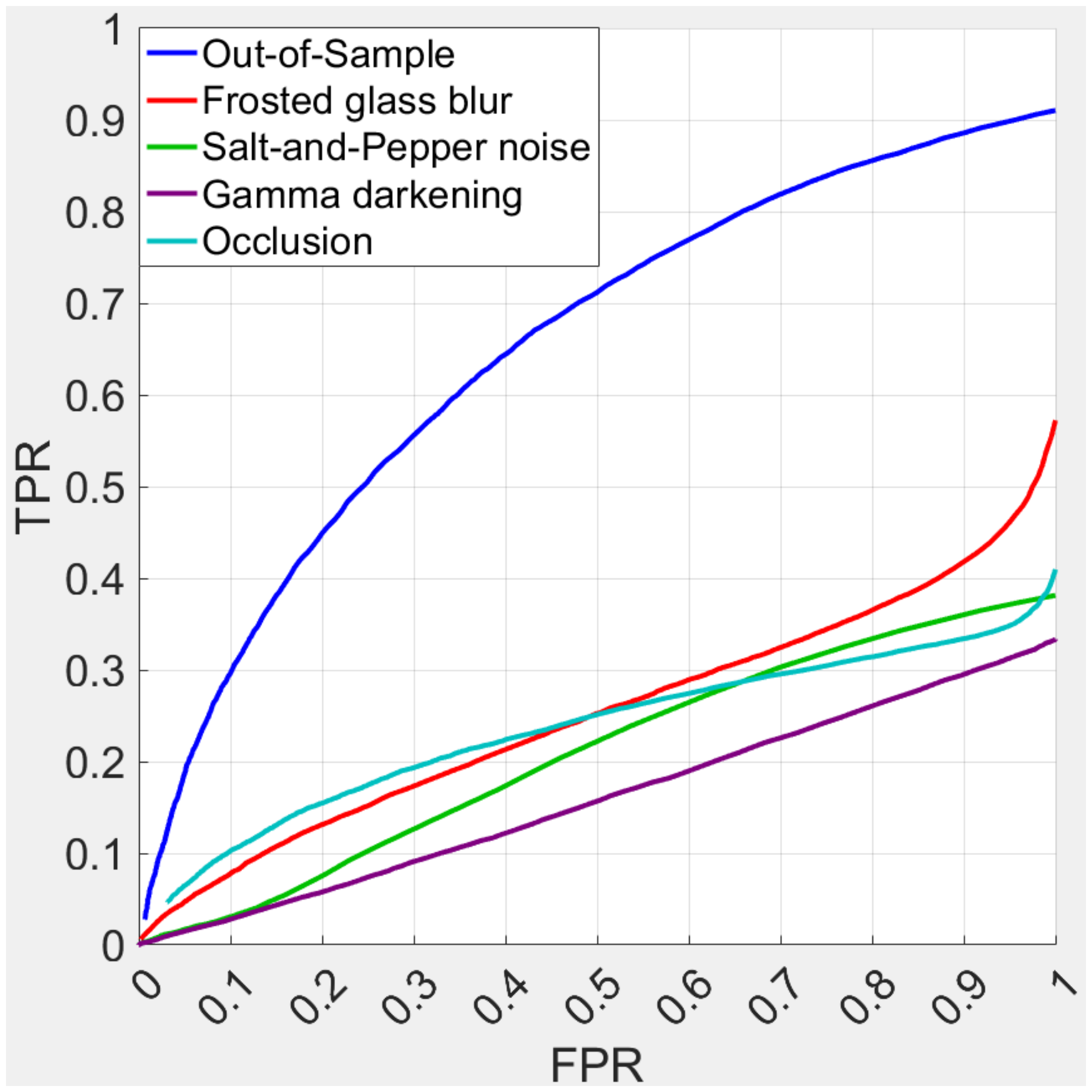}
    \end{center}
    \caption{ROC curves for proposed method in COCO test. The worst performance of each category was chosen to present the tolerance of the algorithm to extreme distortions.}
    \label{figure_3}
\end{figure}

\begin{figure}[!ht]
    \begin{center}
        \includegraphics[width=0.9\linewidth]{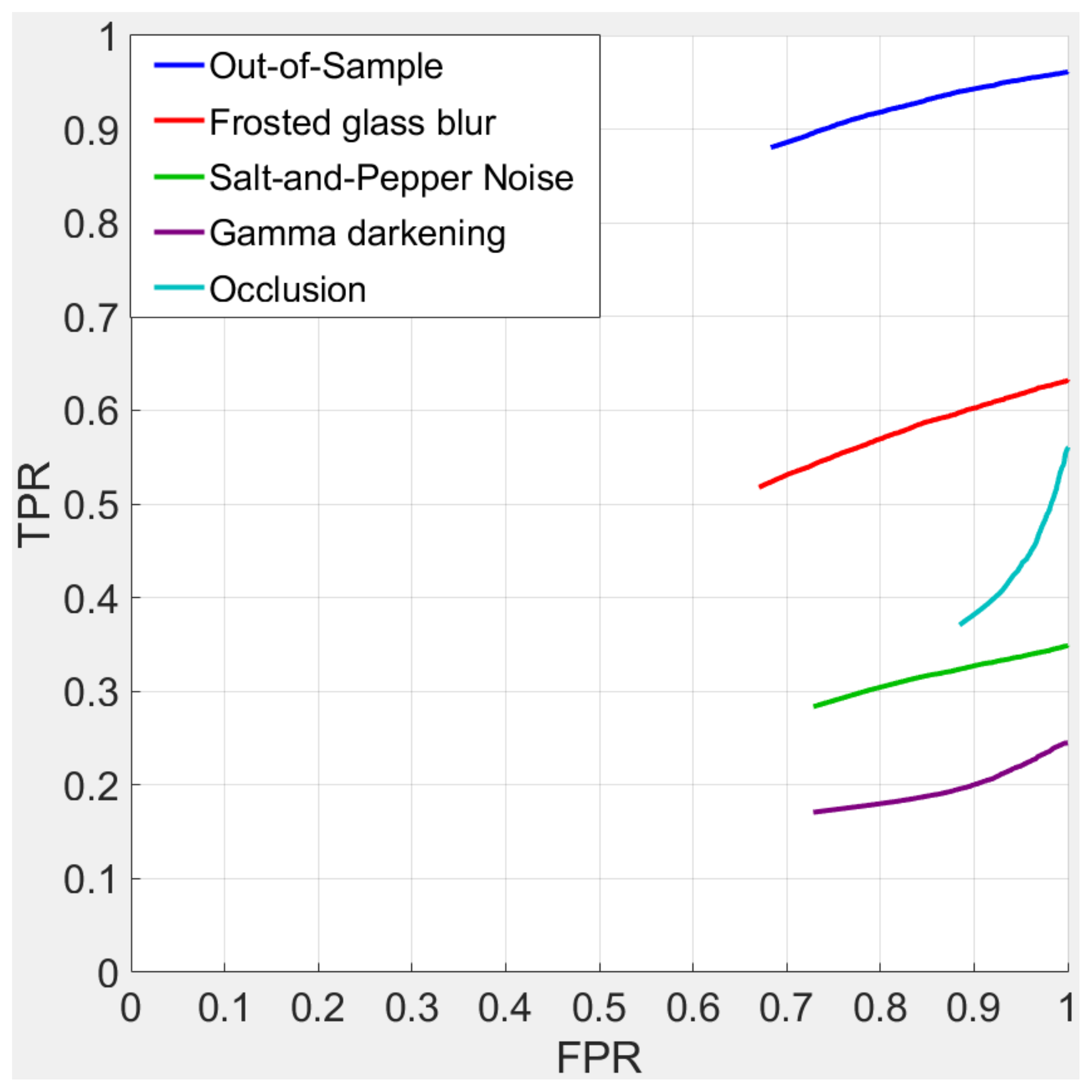}
    \end{center}
    \caption{ROC curves for SR method in COCO test. The worst performance of each category was chosen to present the tolerance of the algorithm to extreme distortions.}
    \label{figure_4}
\end{figure}

After conducting the tests, ROC curves were used to examine the effectiveness of each algorithm. These curves can be seen in Figure \ref{figure_3}-\ref{figure_4}. In general, each point in the ROC curve corresponds to a specific threshold value for the rejection option. If this threshold is set to 0, the algorithm will not reject any input, resulting in a 100\% FPR. The more extreme threshold values will result in lower FPR and True Positive Ratio (TPR) until, at some point, the algorithm rejects all inputs (0\% FPR and TPR). The SR method hits this value when the threshold is set to 1. As the output of Softmax cannot be larger than 1, any output will be rejected. However, since a DNN typically generates high scores for the output, this threshold ends up preventing the SR algorithm from reaching lower FPR rates. On the other hand, the proposed method does not rely on the limit of Softmax output, as it compares the significance of each class to the others. Such a limit will cause a significant gap in AUROC scores, as seen in Table \ref{table_1}.

Judging by the ROC curves, both algorithms start roughly on the same point. This means that both algorithms function similarly when it comes to classification. However, the SR method has the mentioned drawback, which is visible in the curves. 

The comparison must be threshold-independent for it to be fair. Thus, the area under the ROC curve (AUROC) was used as a comparison method. The area calculation must consider the limitations of both algorithms. While the SR algorithm can reach 0\% FPR, it only happens when the threshold is at one (1) or higher, which means the output is not valid. Thus, only the area under the valid parts of the ROC curve was used in calculating the AUROC values. These values can be found in Table \ref{table_1}.

While every distortion reduces the performance, gamma correction has the most significant effect, and blurring has an almost negligible effect on the proposed method. It can be justified by how a classifier works, as changing the intensity of the image makes it harder to separate the objects from the Background class. That being said, the proposed algorithm still outperforms the SR method by a notable margin.

\section{Conclusion}
\label{conclusion}
In this paper, we propose a rejection option for probabilistic classifiers based on Z-test analysis. This method will address the violation of the Closed-World Assumption. By utilizing a probabilistic classifier, each run results in a slightly different class score. A Z-test analyses the mean and standard deviation values for multiple runs to estimate network certainty and filter out uncertain results.

We designed several experiments based on a well-known network configuration (ResNet-18) and datasets (COCO and CIFAR). A comparison with the SR method was made based on AUROC as a threshold-independent metric. The proposed method was shown to have better performance than the SR method by a notable margin while maintaining robustness in the presence of distortions. This makes the proposed method more suitable in safety applications. 

In the future, we will consider expanding the method by merging it with existing tools such as ODIN and covering more complex systems such as object detection.

\bibliographystyle{IEEEbib}
\bibliography{refs}

\end{document}